\documentclass[twoside]{article} 
\usepackage[accepted]{aistats2017}
\usepackage[round]{natbib}
\usepackage{caption}
\usepackage[utf8]{inputenc} 
\usepackage[T1]{fontenc}    
\usepackage{hyperref}       
\usepackage{url}            
\usepackage{booktabs}       
\usepackage{amsfonts}       
\usepackage{nicefrac}       

\usepackage{microtype}      
\usepackage{times}
\usepackage{adjustbox}		
\usepackage{array}		
\usepackage{multirow}           
\usepackage{tabularx}           
\usepackage{booktabs}           
\usepackage{rotating}
\usepackage{wrapfig}
\usepackage{amssymb}
\usepackage{graphicx} 
\usepackage{subfigure} 
\usepackage{tikz}
\usetikzlibrary{shapes.geometric}
\usetikzlibrary{positioning,shapes,shadows,arrows,calc}
\usetikzlibrary{intersections}
\pgfdeclarelayer{background}
\pgfdeclarelayer{foreground}
\pgfsetlayers{background,main,foreground}

\definecolor{lightgray}{gray}{0.5}
\definecolor{lred}{RGB}{255,210,210}
\definecolor{dred}{RGB}{130,0,0} 
\definecolor{dblu}{RGB}{0,0,130}
\definecolor{lblu}{RGB}{210,210,255}
\definecolor{dgre}{RGB}{0,130,0} 
\definecolor{dgra}{RGB}{50,50,50}
\definecolor{mgra}{RGB}{100,100,100}
\definecolor{lgra}{RGB}{220,220,220}
\definecolor{MPG}{RGB}{000,125,122}

\usepackage{algorithm}
\usepackage{algorithmic}

\usepackage{nicefrac}

\usepackage{bm}
\usepackage{amsmath,amsthm,amssymb}
\usepackage{mathtools}

\newcommand{\fabolas}{{\sc{Fabolas}}}

\newcommand{\vx}{{\bm{x}}}

\newcommand{\vy}{{\bm{y}}}

\newcommand{\vlambda}{{\bm{\lambda}}}

\newcommand\transpose{{\textrm{\tiny{\sf{T}}}}}
\newcommand{\argmin}{\operatornamewithlimits{arg\,min}}
\newcommand{\argmax}{\operatornamewithlimits{arg\,max}}

\newcommand{\hide}[1]{}
 
\newcommand{\Exp}{\mathbb{E}}

\newcommand{\note}[1]{
	\noindent~\\
	\vspace{0.25cm}
	\fcolorbox{red}{yellow}{\parbox{0.97\textwidth}{#1\\}}
	\vspace{0.25cm}
}
\renewcommand{\note}[1]{}

\newcommand{\droppedforspace}[1]{}

\begin{document}
\runningtitle{\fabolas: Fast Bayesian Optimization of Machine Learning Hyperparameters on Large Datasets}
\runningauthor{Aaron Klein, Stefan Falkner, Simon Bartels, Philipp Hennig, Frank Hutter}

\twocolumn[

\aistatstitle{Fast Bayesian Optimization of Machine Learning Hyperparameters\\on Large Datasets}

\aistatsauthor{ Aaron Klein$^1$ \And Stefan Falkner$^1$ \And Simon Bartels$^2$ \And Philipp Hennig$^2$ \And Frank Hutter$^1$ }

\aistatsaddress{$^1$\{kleinaa, sfalkner, fh\}@cs.uni-freiburg.de \\  Department of Computer Science \\ University of Freiburg \And $^2$\{simon.bartels, phennig\}@tuebingen.mpg.de \\Department of Empirical Inference \\ Max Planck Institute for Intelligent Systems} ]

\begin{abstract} 
Bayesian optimization has become a successful tool for hyperparameter optimization of machine learning algorithms, such as support vector machines or deep neural networks.
Despite its success, for large datasets, training and validating a single configuration often takes hours, days, or even weeks, which limits the achievable performance.
To accelerate hyperparameter optimization, we propose a generative model for the validation error as a function of training set size, which is learned during the optimization process and allows exploration of preliminary configurations on small subsets, by extrapolating to the full dataset.
We construct a Bayesian optimization procedure, dubbed \fabolas{}, which models loss and training time as a function of dataset size and automatically trades off high information gain about the global optimum against computational cost.
Experiments optimizing support vector machines and deep neural networks show that \fabolas{} often finds high-quality solutions 10 to 100 times faster than other state-of-the-art Bayesian optimization methods or the recently proposed bandit strategy Hyperband.
\end{abstract}

\section{Introduction}\label{sec:intro}

The performance of many machine learning algorithms hinges on certain hyperparameters.
For example, the prediction error of non-linear support vector machines depends on regularization and kernel hyperparameters $C$ and $\gamma$; 
and modern neural networks are sensitive to a wide range of hyperparameters, including learning rates, momentum terms, number of units per layer, dropout rates, weight decay, etc.~\citep{montavon-book12a}.
The poor scaling of na\"ive methods like grid search with dimensionality has driven interest in more sophisticated hyperparameter optimization methods over the past years~\citep{bergstra-nips11a,hutter-lion11a,bergstra-jmlr12a,snoek-nips12a,bardenet-icml13a,bergstra-icml13a,swersky-nips13,swersky-corr14,snoek-icml14a,snoek-icml15a}.
\emph{Bayesian optimization} has emerged as an efficient framework, achieving impressive successes.
For example, in several studies, it found better instantiations of convolutional network hyperparameters than domain experts, repeatedly improving the top score on the CIFAR-10~\citep{krizhevsky-tech09a} benchmark without data augmentation~\citep{snoek-nips12a,domhan-ijcai15,snoek-icml15a}.

In the traditional setting of Bayesian hyperparameter optimization, the loss of a machine learning algorithm with hyperparameters $\vx \in \mathbb{X}$ is treated as the ``black-box'' problem of finding $ \argmin_{\vx \in \mathbb{X}} f(\vx)$, where the only mode of interaction with the objective $f$ is to evaluate it for inputs $\vx \in \mathbb{X}$.
If individual evaluations of $f$ on the entire dataset require days or weeks, only very few evaluations are possible, limiting the quality of the best found value. 
Human experts instead often study performance on subsets of the data first, to become familiar with its characteristics before gradually increasing the subset size~\citep{bot-nn12a,montavon-book12a}.
This approach can still outperform contemporary Bayesian optimization methods.

Motivated by the experts' strategy, here we leverage dataset size as an additional degree of freedom enriching the representation of the optimization problem.
We treat the size of a randomly subsampled dataset $N_{sub}$ as an additional input to the blackbox function, and allow the optimizer to actively choose it at each function evaluation.
This allows Bayesian optimization to mimic and improve upon human experts when exploring the hyperparameter space.
In the end, $N_{sub}$ is not a hyperparameter itself, but the goal remains a good performance on the full dataset, i.e.~$N_{sub} = N$.

Hyperparameter optimization for large datasets has been explored by other authors before.
Our approach is similar to Multi-Task Bayesian optimization by \citet{swersky-nips13}, where knowledge is transferred between a finite number of correlated tasks.
If these tasks represent manually-chosen subset-sizes, this method also tries to find the best configuration for the full dataset by evaluating smaller, cheaper subsets.
However, the discrete nature of tasks in that approach requires evaluations on the entire dataset to learn the necessary correlations.
Instead, our approach exploits the regularity of performance across dataset size, enabling generalization to the full dataset without evaluating it directly.

Other approaches for hyperparameter optimization on large datasets include work by \citet{nickson-archive}, who estimated a configuration's performance on a large dataset by evaluating several training runs on small, random subsets of fixed, manually-chosen sizes.
\citet{krueger-jmlr15a} showed that, in practical applications, small subsets can suffice to estimate a configuration's quality, and proposed a cross-validation scheme that sequentially tests a fixed set of configurations on a growing subset of the data, discarding poorly-performing configurations early.

In parallel work\footnote{Hyperband was first described in a 2016 arXiv paper~\citep{li-arxiv16}, and \fabolas{} was first described in a 2015 NIPS workshop paper~\citep{klein-bayesopt15}}, \citet{li-iclr17} proposed a multi-arm bandit strategy, called Hyperband, which dynamically allocates more and more resources to randomly sampled configurations based on their performance on subsets of the data.
Hyperband assures that only well-performing configurations are trained on the full dataset while discarding bad ones early.
Despite its simplicity, in their experiments the method was able to outperform well-established Bayesian optimization algorithms.

In \textsection\ref{sec:bo}, we review Bayesian optimization, in particular the Entropy Search algorithm and the related method of Multi-Task Bayesian optimization.
In \textsection\ref{sec:fabolas}, we introduce our new Bayesian optimization method \fabolas{} for hyperparameter optimization on large datasets.
In each iteration, \fabolas{} chooses the configuration $x$ and	dataset size $N_{sub}$ predicted to yield most information about the loss-minimizing configuration on the full dataset per \emph{unit time spent}.
In \textsection\ref{sec:experiments}, a broad range of experiments with support vector machines and various deep neural networks show \fabolas{} often identifies good hyperparameter settings 10 to 100 times faster than state-of-the-art Bayesian optimization methods acting on the full dataset as well as Hyperband.

\section{Bayesian optimization}\label{sec:bo}

Given a black-box function $f: \mathbb{X} \rightarrow \mathbb{R}$, Bayesian optimization\footnote{Comprehensive tutorials are presented by \citet{brochu-corr10a} and \citet{shahriari-ieee16}.} aims to find an input $\vx_{\star} \in \argmin_{\vx \in \mathbb{X}} f(\vx)$ that globally minimizes $f$.
It requires a prior $p(f)$ over the function and an acquisition function $a_{p(f)}: \mathbb{X} \rightarrow \mathbb{R}$ quantifying the {\it utility} of an evaluation at any $\vx$.
With these ingredients, the following three steps are iterated \citep{brochu-corr10a}: (1)~find the most promising \mbox{$\vx_{n+1} \in \argmax{a_p(\vx)}$} by numerical optimization; (2)~evaluate the expensive and often noisy function $y_{n+1} \sim f(\vx_{n+1}) + \mathcal{N}(0,\sigma^2)$ and add the resulting data point $(\vx_{n+1},y_{n+1})$ to the set of observations $\mathcal{D}_n=(\vx_j,y_j)_{j=1...n}$; and (3)~update $p(f\mid\mathcal{D}_{n+1})$ and $a_{p(f\mid\mathcal{D}_{n+1})}$.
Typically, evaluations of the acquisition function $a$ are cheap compared to evaluations of $f$ such that the optimization effort is negligible.

\subsection{Gaussian Processes}
\label{sec:gp}

Gaussian processes (GP) are a prominent choice for $p(f)$, thanks to their descriptive power and analytic tractability \citep[e.g.][]{rasmussen-book06a}.
Formally, a GP is a collection of random variables, such that every finite subset of them follows a multivariate normal distribution.
A GP is identified by a mean function $m$ (often set to $m(\vx)=0\; \forall \vx\in\mathbb{X}$), and a positive definite covariance function (kernel) $k$.
Given observations $\mathcal{D}_n=(\vx_j,y_j)_{j=1...n}=(\boldsymbol{X},\vy)$ with joint Gaussian likelihood $p(\vy\mid \boldsymbol{X}, f(\boldsymbol{X}))$, the posterior $p(f|\mathcal{D}_n)$ follows another GP, with mean and covariance functions of tractable, analytic form.

The covariance function determines how observations influence the prediction.
For the hyperparameters we wish to optimize, we adopt the Mat\'ern $\nicefrac{5}{2}$ kernel \citep{matern1960spatial}, in its Automatic Relevance Determination form \citep{ARD}.
This stationary, twice-differentiable model constitutes a relatively standard choice in the Bayesian optimization literature.
In contrast to the Gaussian kernel popular elsewhere, it makes less restrictive smoothness assumptions, which can be helpful in the optimization setting \citep{snoek-nips12a}:
\begin{equation}
\begin{split}
\label{eq:matern}
k_{\nicefrac 52}(\vx,\vx^{\prime})&=\theta\left(1+\sqrt{5}d_\vlambda(\vx,\vx')\right.\\ 
& \left. +\nicefrac{5}{3}d^2_\vlambda(\vx,\vx')\right)e^{-\sqrt{5}d_\vlambda(\vx,\vx')}.
\end{split}
\end{equation}
Here, $\theta$ and $\vlambda$ are free parameters---hyperparameters of the GP surrogate model---and $d_\vlambda(\vx,\vx')=(\vx-\vx')^\transpose\operatorname{diag}(\vlambda)(\vx-\vx')$ is the Mahalanobis distance.
For the dataset size dependent performance and cost, we construct a custom kernel in \ref{sec:kernels}.
An additional hyperparameter of the GP model is a overall noise covariance needed to handle noisy observations.
For clarity: These GP hyperparameters are \emph{internal} hyperparameters of the Bayesian optimizer, as opposed to those of the target machine learning algorithm to be tuned. Section \ref{sec:implementation} shows how we handle them.

\subsection{Acquisition functions}
\label{sec:acquisition_functions}

The role of the acquisition function is to trade off exploration vs.\ exploitation. Popular choices include Expected Improvement (EI) \citep{mockus-tgo78a}, Upper Confidence Bound (UCB)~\citep{srninivas-icml10a}, Entropy Search (ES)~\citep{hennig-jmlr12a}, and Predictive Entropy Search (PES)~\citep{hernandez-nips14}.
In our experiments, we will use EI and ES.

We found EI to perform robustly in most applications, providing a solid baseline;
it is defined as
\begin{align}
\label{eq:ei}
a_{\text{EI}}(\vx|\mathcal{D}_n)&=\Exp_{p}[\max(f_{\text{min}}-f(\vx), 0)]\,.
\end{align}
where $f_{\text{min}}$ is the best function value known (also called the \emph{incumbent}).
This expected drop over the best known value is high for points predicted to have small mean and/or large variance.

ES is a more recent acquisition function that selects evaluation points based on the predicted \emph{information gain} about the optimum, rather than aiming to evaluate near the optimum.
At the heart of ES lies the probability distribution $p_{\min}(\vx \mid \mathcal{D}):=p(\vx \in \argmin_{\vx' \in \mathbb{X}} f(\vx') \mid \mathcal{D})$, the belief about the function's minimum given the prior on $f$ and observations $\mathcal{D}$.
The \emph{information gain} at $\vx$ is then measured by the expected Kullback-Leibler divergence (relative entropy) between $p_{\min}(\cdot \mid \mathcal{D} \cup \{(\vx, y)\})$ and the uniform distribution $u(\vx)$, with expectations taken over the measurement $y$ to be obtained at $\vx$: 
\begin{equation}
\begin{split}
\label{eq:es}
 a_{\text{ES}}(\vx):&=\Exp_{p(y\mid \vx, \mathcal{D})}\left[ \int p_{\min}(\vx' \mid \mathcal{D} \cup \{(\vx, y)\} ) \right.\\
 & \left. \cdot \log\frac{p_{\min}(\vx' \mid \mathcal{D} \cup \{(\vx, y)\})}{u(\vx')}\ \mathrm{d}\vx'\right].
\end{split}
\end{equation}
The primary numerical challenge in this framework is the computation of $p_{\min}(\cdot \mid \mathcal{D} \cup \{(\vx, y)\})$ and the integral above.
Due to the intractability, several approximations have to be made.
We refer to \citet{hennig-jmlr12a} for details, as well as to the supplemental material (Section A), where we also provide pseudocode for our implementation.
Despite the conceptual and computational complexity of ES, it offers a well-defined concept for information gained from function evaluations, which can be meaningfully traded off against other quantities, such as the evaluations' cost. 

PES refers to the same acquisition function, but uses different approximations to compute it.
In Section \ref{sec:implementation} we describe why, for our application, ES was the more direct choice.

\subsection{Multi-Task Bayesian optimization}\label{sec:multibo}

The \emph{Multi-Task} Bayesian optimization (MTBO) method of \citet{swersky-nips13} refers to a general framework for optimizing in the presents of different, but correlated tasks.
Given a set of such tasks $\mathbb{T} = \{1, \ldots, T\}$, the objective function $f: \mathbb{X}\times\mathbb{T} \rightarrow \mathbb{R}$ corresponds to evaluating a given $\vx \in \mathbb{X}$ on one of the tasks $t \in \mathbb{T}$.
The relation between points in $\mathbb{X}\times\mathbb{T}$ is modeled via a GP using a product kernel:
\begin{align}
k_{\text{MT}}((\vx, t), (\vx', t^{\prime})) = k_T(t, t^{\prime}) \cdot k_{\nicefrac 52}(\vx, \vx')\,.
\end{align}
The kernel $k_T$ is represented implicitly by the Cholesky decomposition of $k(\mathbb{T},\mathbb{T})$ whose entries are sampled via MCMC together with the other hyperparameters of the GP.
By considering the distribution over the optimum on the target task $t_*\in \mathbb{T}$, $p_{\min}^{t_*}(\vx \mid \mathcal{D}):=p(\vx \in \argmin_{\vx' \in \mathbb{X}} f(\vx', t=t_*) \mid \mathcal{D})$, and computing any information w.r.t.\ it, \citet{swersky-nips13} use the information gain per unit cost as their acquisition function\footnote{In fact, \citet{swersky-nips13} deviated slightly from this formula (which follows the ES approach of \cite{hennig-jmlr12a}) by considering the difference in information gains in $p_{\min}^{t_*}(\vx \mid \mathcal{D})$ and $p_{\min}^{t_*}(\vx \mid \mathcal{D} \cup \{(\vx, y)\})$. They stated this to work better in practice, but we did not find evidence for this in our experiments and thus, for consistency, use the variant presented here throughout.}:
\begin{align}
\label{eq:mtbo_acq}
\nonumber
a_{\text{MT}}(\vx, t):&=\frac{1}{c(\vx,t)}\Exp_{p(y\mid \vx, t, \mathcal{D})}\left[ \int p_{\min}^{t_*}(\vx' \mid \mathcal{D}' ) \right. \\
& \left. \cdot\log\frac{p_{\min}^{t_*}(\vx' \mid \mathcal{D}')}{u(\vx')}\ \mathrm{d}\vx'\right]\,,
\end{align}
where $\mathcal{D}' = \mathcal{D} \cup \{(\vx, t, y)\}$.
The expectation represents the information gain on the target task averaged over the possible outcomes of $f(\vx, t)$ based on the current model. 
If the cost $c(\vx, t)$ of a configuration $\vx$ on task $t$ is not known a priori it can be modelled the same way as the objective function.

This model supports machine learning hyperparameter optimization for large datasets by using discrete dataset sizes as tasks. \citet{swersky-nips13} indeed studied this  approach for the special case of $\mathbb{T} = \{0,1\}$, representing a small and a large dataset; this will be a baseline in our experiments.

\section{Fast Bayesian optimization for large datasets}\label{sec:fabolas}

Here, we introduce our new approach for FAst Bayesian Optimization on LArge data Sets (\fabolas{}).
While traditional Bayesian hyperparameter optimizers model the loss of machine learning algorithms on a given dataset as a blackbox function $f$ to be minimized,
\fabolas{} models loss and computational cost \emph{across dataset size} and uses these models to carry out Bayesian optimization with an extra degree of freedom.
The blackbox function $f: \mathbb{X} \times \mathbb{R} \rightarrow \mathbb{R}$ now takes another input representing the data subset size; we will use relative sizes $s = N_{sub}/N \in [0,1]$, with $s=1$ representing the entire dataset.
While the eventual goal is to minimize the loss $f(\vx,s=1)$ for the entire dataset, evaluating $f$ for smaller $s$ is usually cheaper, and the function values obtained correlate across $s$.
Unfortunately, this correlation structure is initially unknown, so the challenge is to design a strategy that trades off the cost of function evaluations against the benefit of learning about the scaling behavior of $f$ and, ultimately, about which configurations work best on the full dataset.
Following the nomenclature of \citet{williams}, we call $s\in [0,1]$ an \emph{environmental variable} that can be changed freely \emph{during} optimization, but that is set to $s=1$ (i.e., the entire dataset size), at evaluation time.

We propose a principled rule for the automatic selection of the next $(\vx,s)$ pair to evaluate.
In a nutshell, where standard Bayesian optimization would always run configurations on the full dataset, we use ES to reason about, how much can be learned about performance on the full dataset from an evaluation at any $s$.
In doing so, \fabolas{} automatically determines the amount of data necessary to (usefully) extrapolate to the full dataset.

\begin{figure*}[t]
 \centering
 \begin{subfigure}[$ s = \nicefrac{1}{128}$]{\includegraphics[height=3.05cm,width=0.24\textwidth]{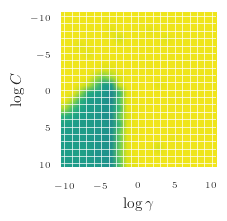}} \end{subfigure}
 \begin{subfigure}[$ s = \nicefrac{1}{16}$]{\includegraphics[height=3.05cm,width=0.24\textwidth]{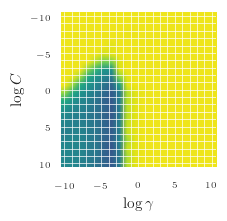}} \end{subfigure}
 \begin{subfigure}[$ s = \nicefrac{1}{4}$]{\includegraphics[height=3.05cm,width=0.24\textwidth]{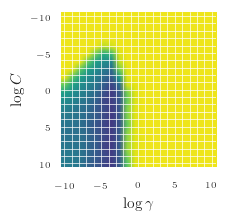}} \end{subfigure}
 \begin{subfigure}[$ s = 1$]{\includegraphics[height=3.15cm,width=0.24\textwidth]{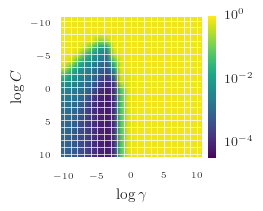}} \end{subfigure}
\caption{Validation error of a grid of $400$ SVM configurations (20 settings of each of the regularization parameter $C$ and kernel parameter $\gamma$, both on a log-scale in $[-10, 10]$) for subsets of the MNIST dataset~\citep{lecun-isp01a} of various sizes $N_{\text{sub}}$.
 Small subsets are quite representative: The validation error of bad configuration (yellow) remains constant at around $0.9$, whereas the region of good configurations (blue) does not change drastically with $s$.}
\label{fig:grid}
\end{figure*}

For an initial intuition on how performance changes with dataset size, we evaluated a grid of $400$ configurations of a support vector machine (SVM) on subsets of the MNIST dataset~\citep{lecun-isp01a} ; MNIST has $N=50\,000$ data points and we evaluated relative subset sizes $s \in \{\nicefrac{1}{512}, \nicefrac{1}{256}, \nicefrac{1}{128}, \ldots , \nicefrac{1}{4},\nicefrac{1}{2}, 1\}$.
Figure \ref{fig:grid} visualizes the validation error of these configurations on $s=\nicefrac{1}{128}$, $\nicefrac{1}{16}$, $\nicefrac{1}{4}$, and $1$.
Evidently, just $\nicefrac{1}{128}$ of the dataset is quite representative and sufficient to locate a reasonable configuration.
Additionally, there are no deceiving local optima on smaller subsets.
Based on these observations, we expect that relatively small fractions of the dataset yield representative performances and therefore vary our relative size parameter $s$ on a logarithmic scale.

\subsection{Kernels for loss and computational cost}\label{sec:kernels}

To transfer the insights from this illustrative example into a formal model for the loss and cost across subset sizes, we extend the GP model by an additional input dimension, namely $s \in [0,1]$.
This allows the surrogate to extrapolate to the full data set at $s=1$ without necessarily evaluating there.
We chose a factorized kernel, consisting of the standard stationary kernel over hyperparameters, multiplied with a finite-rank (``degenerate'') covariance function in $s$:
\begin{equation}
 k\left( (\vx,s), (\vx', s') \right) = k_{\nicefrac 52} \left( \vx, \vx' \right) \cdot \left(  \phi^T(s) \cdot \Sigma_{\phi} \cdot \phi(s') \right)\,.
 \label{eq:our_kernel}
\end{equation}
Since any choice of the basis function $\phi$ yields a positive semi-definite covariance function, this provides a flexible language for prior knowledge relating to $s$.
We use the same form of kernel to model the loss $f$ and cost $c$, respectively, but with different basis functions $\phi_{f}$ and $\phi_{c}$.

The loss of a machine learning algorithms usually decreases with more training data. We incorporate this behavior by choosing $\phi_{f}(s) = (1, (1 - s)^2)^T$ to enforce monotonic predictions with an extremum at $s=1$. This kernel choice is equivalent to Bayesian linear regression with these basis functions and Gaussian priors on the weights. 

To model computational cost $c$, we note that the complexity usually grows with relative dataset size $s$. To fit polynomial complexity $\mathcal O(s^\alpha)$ for arbitrary $\alpha$ and simultaneously enforce positive predictions, we model the log-cost and use $\phi_c(s) = (1,s)^T$.
As above, this amounts to Bayesian linear regression with shown basis functions.

In the supplemental material (Section B), we visualize scaling of loss and cost with $s$ for the SVM example above and show that our kernels indeed fit them well. We also evaluate the possibility of modelling the heteroscedastic noise introduced by subsampling the data (supplementary material, Section C).


\subsection{Formal algorithm description}

\fabolas{} starts with an initial design, described in more detail in Section \ref{sec:initial_design}.
Afterwards, at the beginning of each iteration it fits GPs for loss and computational cost across dataset sizes $s$ using the kernel from Eq.~\ref{eq:our_kernel}. 
Then, capturing the distribution of the optimum for $s=1$ using $p_{\min}^{s=1}(\vx \mid \mathcal{D}):=p(\vx \in \argmin_{\vx' \in \mathbb{X}} f(\vx', s=1) \mid \mathcal{D})$, it selects the maximizer of the following acquisition function to trade off information gain versus cost:
\begin{align}
\nonumber
a_{\text{F}}(\vx, s):&=\frac{1}{c(\vx,s) + c_{\text{overhead}}} \\
\label{eq:fabolas_acq}
& \Exp_{p(y\mid \vx, s, \mathcal{D})}\left[ \int p_{\min}^{s=1}(\vx' \mid \mathcal{D} \cup \{(\vx, s, y)\} )\cdot \right.\\
\nonumber
& \left. \log\frac{p_{\min}^{s=1}(\vx' \mid \mathcal{D} \cup \{(\vx, s, y)\})}{u(\vx')}\ \mathrm{d}\vx'\right].
\end{align}
Algorithm \ref{alg:fabolas} shows pseudocode for \fabolas{}. We also provide an open-source implementation at \href{https://github.com/automl/RoBO}{https://github.com/automl/RoBO}. 

\begin{algorithm}
\caption{Fast BO for Large Datasets (\fabolas{})}
\label{alg:fabolas}
\begin{algorithmic}[1]
{
\STATE Initialize data $\mathcal{D}_{0}$ using an initial design.\label{line:initial_design}
\FOR{$t=1,2,\dots$}
  \STATE Fit GP models for $f(\vx, s)$ and $c(\vx, s)$ on data $\mathcal{D}_{t-1}$\label{line:fit_models}
  \STATE Choose $(\vx_{t},s_t)$ by maximizing the acquisition function in Equation \ref{eq:fabolas_acq}.\label{line:opt_acq}
  \STATE Evaluate $y_t\sim f(\vx_{t},s_t) + \mathcal{N}(0,\sigma^2)$, also measuring cost $z_t\sim c(\vx_{t},s_t) + \mathcal{N}(0,\sigma^2_{c})$, and augment the data: $\mathcal{D}_{t} = \mathcal{D}_{t-1} \cup \{(\vx_{t}, s_t, y_t, z_t)\}$
  \STATE Choose incumbent $\hat{\vx}_t$ based on the predicted loss at $s=1$ of all $\{\vx_1, \vx_2, \dots, \vx_t\}$.\label{line:incumbent}
\ENDFOR
}
\end{algorithmic}
\end{algorithm}

Our proposed acquisition function resembles the one used by MTBO (Eq.~\ref{eq:mtbo_acq}), with two differences: First, MTBO's discrete tasks $t$ are replaced by a continuous dataset size $s$ (allowing to learn correlations without evaluations at $s=1$, and to choose the appropriate subset size automatically).
Second, the prediction of computational cost is augmented by the overhead of the Bayesian optimization method.
This inclusion of the reasoning overhead is important to appropriately reflect the information gain per unit time spent: it does not matter whether the time is spent with a function evaluation or with reasoning about which evaluation to perform.
In practice, due to cubic scaling in the number of data points of GPs and the computational complexity of approximating $p_{\min}^{s=1}$, the additional overhead of \fabolas{} is within the order of minutes, such that differences in computational cost in the order of seconds become negligible in comparison.\footnote{The same is true for standard ES and MTBO, but was never exploited as no emphasis was put on the total wall clock time spent for the hyperparameter optimization.
We want to emphasize that we express budgets in terms of wall clock time (not function evaluations) since this is natural in most practical applications.}

Being an anytime algorithm, \fabolas{} keeps track of its incumbent at each time step.
To select a configuration that performs well on the full dataset, it predicts the loss of all evaluated configurations at $s=1$ using the GP model and picks the minimizer.
We found this to work more robustly than globally minimizing the posterior mean, or similar approaches.

\subsection{Initial design}\label{sec:initial_design}

It is common in Bayesian optimization to start with an initial design of points chosen at random or from a Latin hypercube design to allow for reasonable GP models as starting points.
To fully leverage the speedups we can obtain from evaluating small datasets, we bias this selection towards points with small (cheap) datasets in order to improve the prediction for dependencies on $s$:
We draw $k$ random points in $\mathbb{X}$ ($k=10$ in our experiments) and evaluate them on different subsets of the data (for instance on the support vector machine experiments we used $s \in \{\nicefrac{1}{64}, \nicefrac{1}{32}, \nicefrac{1}{16}, \nicefrac{1}{8}\}$).
This provides information on scaling behavior, and, assuming that costs increase linearly or superlinearly with $s$, these $k$ function evaluations cost less than $\frac{k}{8}$ function evaluations on the full dataset. This is important as the cost of the initial design, of course, counts towards \fabolas{}' runtime. 

\subsection{Implementation details}\label{sec:implementation}
The presentation of \fabolas{} above omits some details that impact the performance of our method. 
As it has become standard in Bayesian optimization \citep{snoek-nips12a}, we use Markov-Chain Monte Carlo (MCMC) integration to marginalize over the GPs hyperparameters (we use the emcee package \citep{Foreman}).
To accelerate the optimization, we use hyper-priors to emphasize meaningful values for the parameters,
chiefly adopting the choices of the {\sc spearmint} toolbox \citep{snoek-nips12a}:
a uniform prior between $[-10, 2]$ for all length scales $\vlambda$ in log space, a lognormal prior ($\mu_a = 0$, $\sigma_a^2=1$) for the covariance amplitude $\theta$, and a horseshoe prior with length scale of $0.1$ for the noise variance $\sigma^2$.


We used the original formulation of ES by \citet{hennig-jmlr12a} rather than the recent reformulation of PES by \citet{hernandez-nips14}.
The main reason for this is that the latter prohibits non-stationary kernels due to its use of Bochner's theorem for a spectral approximation.
PES could in principle be extended to work for our particular choice of kernels (using an Eigen-expansion, from which we could sample features); since this would complicate making modifications to our kernel, we leave it as an avenue for future work, but note that in any case it may only further improve our method.
To maximize the acquisition function we used the blackbox optimizer DIRECT \citep{jones-jgo01a} and CMAES \citep{hansen-eda06}. 

\section{Experiments}\label{sec:experiments}

\begin{figure*}[t]
 \centering
 \begin{subfigure}
  \centering
  \includegraphics[width=0.317\textwidth]{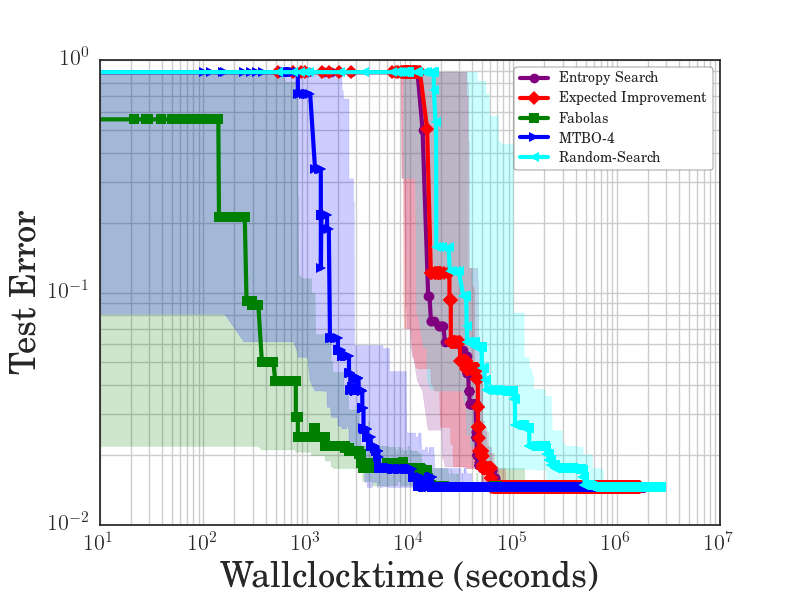}
  \includegraphics[width=0.317\textwidth]{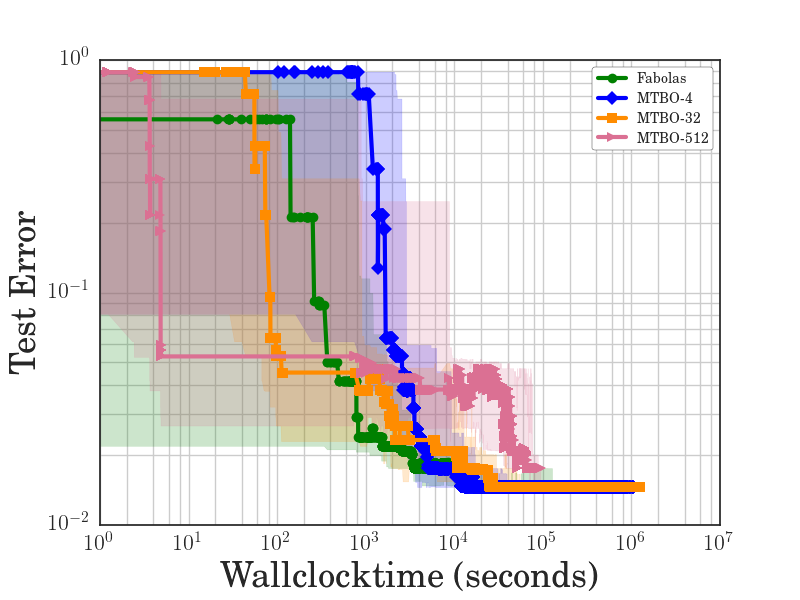}
  \includegraphics[width=0.35\textwidth]{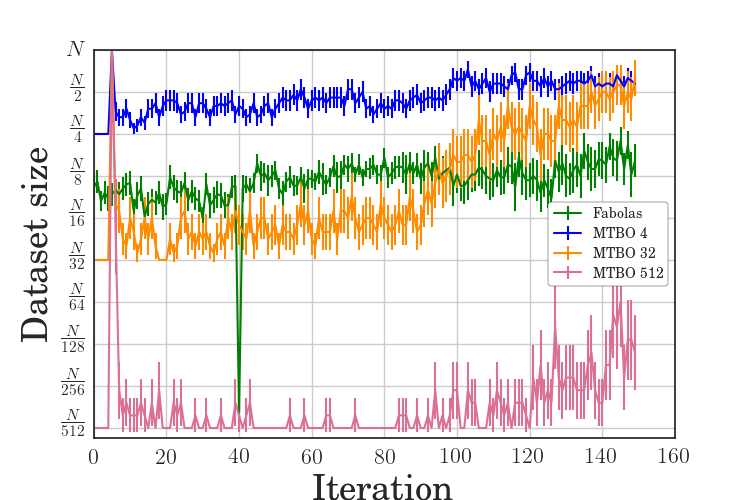}
 \end{subfigure}
 \caption{Evaluation on SVM grid on MNIST. (Left) Baseline comparison of test performance of the methods' selected incumbents over time. 
 (Middle) Test performance over time for variants of MTBO with different dataset sizes for the auxiliary task. 
 (Right) Dataset size \fabolas{} and MTBO pick in each iteration to trade off small cost and high information gain; unlike elsewhere in the paper, this right plot shows mean $\pm \nicefrac{1}{4}$ stddev of $30$ runs (medians would only take two values for MTBO).
}
\label{fig:svm_on_grid}
\end{figure*}

For our empirical evaluation of \fabolas{}, we compared it to standard Bayesian optimization (using EI and ES as acquisition functions), MTBO, and Hyperband.
For each method, we tracked wall clock time (counting both optimization overhead and the cost of function evaluations, including the initial design), storing the incumbent returned after every iteration.
In an offline validation step, we then trained models with all incumbents on the full dataset and measured their test error. We plot these test errors throughout.\footnote{The  residual network in Section \ref{sec:resnet_exp} is an exception: here, we trained networks with the incumbents on the full training set (50000 data points, augmented to 100000 as in the original code) and then measured and plotted performance on the validation set.} 
To obtain error bars, we performed 10 independent runs of each method with different seeds (except on the grid experiment, where we could afford 30 runs per method) and plot medians, along with ${25}^{\text{th}}$ and $75^{\text{th}}$ percentiles for all experiments.
Details on the hyperparameter ranges used in every experiment are given in the supplemental material (Section D).

We implemented Hyperband following \citet{li-iclr17} using the recommended setting for the parameter $\eta = 3$ that controls the intermediate subset sizes.
For each experiment, we adjusted the budget allocated to each Hyperband iteration to allow the same minimum dataset size as for \fabolas{}: 10 times the number of classes for the support vector machine benchmarks and the maximum batch size for the neural network benchmarks.
We also followed the prescribed incumbent estimation after each iteration as the configuration with the best performance on the full dataset size.

\subsection{Support vector machine grid on MNIST}\label{sec:svm}

First, we considered a benchmark allowing the comparison of the various Bayesian optimization methods on ground truth: our SVM grid on MNIST (described in Section \ref{sec:fabolas}), for which we had performed all function evaluations beforehand, measuring loss and cost 10 times for each configuration $\vx$ and subset size $s$ to account for performance variations.
(In this case, we computed each method's wall clock time in each iteration as its summed optimization overheads so far, plus the summed costs for the function values it queried so far.)

MTBO requires choosing the number of data points in its auxiliary task. 
Figure \ref{fig:svm_on_grid} (middle) evaluates MTBO variants with a single auxiliary task with a relative size of $\nicefrac{1}{4}$, $\nicefrac{1}{32}$, and $\nicefrac{1}{512}$, respectively. 
With auxiliary tasks at either $s=\nicefrac{1}{512}$ or $\nicefrac{1}{32}$, MTBO improved quickly, but converged more slowly to the optimum; we believe small correlations between the tasks cause this.
Figure \ref{fig:svm_on_grid} (right) shows the dataset sizes chosen by the different algorithms during the optimization; all methods slowly increased the average subset size used over time.
An auxiliary task with $s=\nicefrac{1}{4}$ worked best and we used this for MTBO in the remaining experiments.

At first glance, one might expect many tasks (e.g., with a task for each $s \in \{\nicefrac{1}{512}, \nicefrac{1}{256}, \dots, \nicefrac{1}{2}, 1\} $) to work best, but quite the opposite is true.
In preliminary experiments, we evaluated MTBO with up to 3 auxiliary tasks ($s=\nicefrac{1}{4}$, $\nicefrac{1}{32}$, and $\nicefrac{1}{512}$), but found performance to strongly degrade with a growing number of tasks.
We suspect that the ${\vert T\vert }\choose{2}$ kernel parameters that have to be learned for the discrete task kernel for $\vert T \vert$ tasks are the main reason.
If the MCMC sampling is too short, the correlations are not appropriately reflected, especially in early iterations, and an adjusted sampling creates a large computational overhead that dominates wall-clock time.
We therefore obtained best performance with only one auxiliary task.

Figure \ref{fig:svm_on_grid} (left) shows results using EI, ES, random search, MTBO and {\sc{fabolas}} on this SVM benchmark. 
EI and ES perform equally well and find the best configuration (which yields an error of $0.014$, or $1.4\%$) after around $10^5$ seconds, roughly five times faster than random search.
MTBO achieves good performance faster, requiring only around $2\times 10^4$ seconds to find the global optimum.
\fabolas{} is roughly another order of magnitude faster than MTBO in finding good configurations, and finds the global optimum at the same time.

\begin{figure*}[t]
 \centering
 \begin{subfigure}
  \centering
  \includegraphics[width=0.327\textwidth]{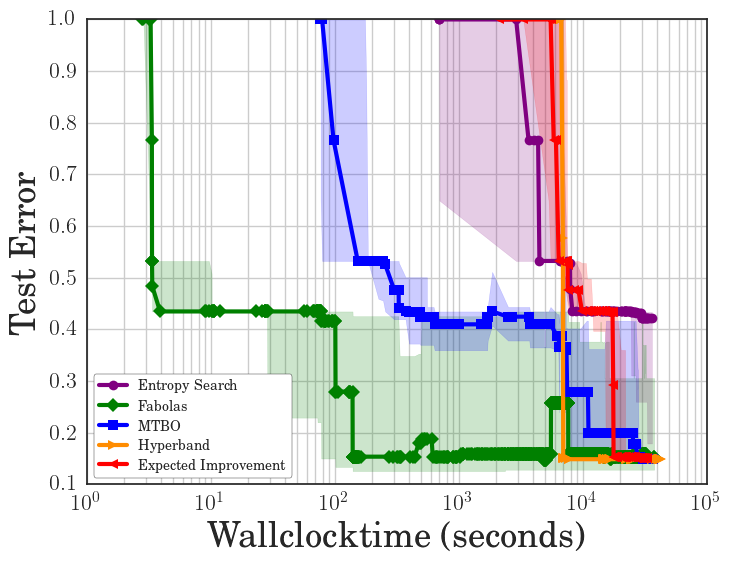}
  \includegraphics[width=0.327\textwidth]{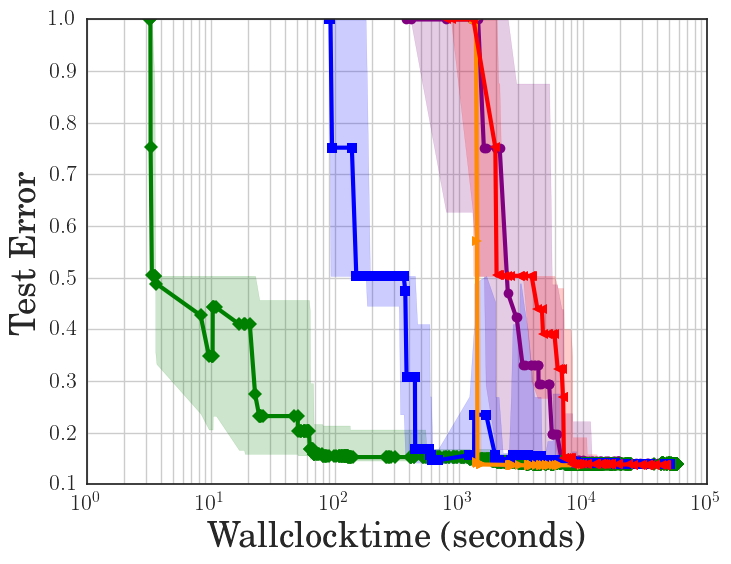}
  \includegraphics[width=0.327\textwidth]{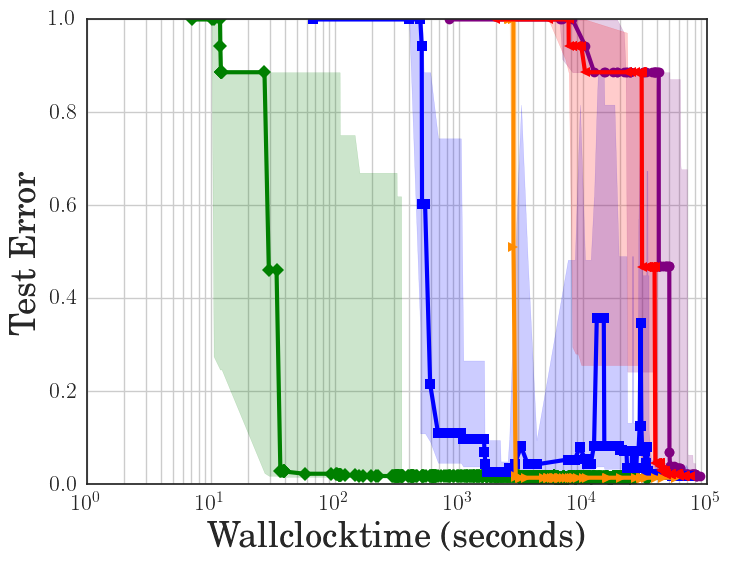}
 \end{subfigure}
\caption{SVM hyperparameter optimization on the datasets covertype (left), vehicle (middle) and MNIST(right). At each time, the plots show test performance of the methods' respective incumbents. \fabolas{} finds a good configuration between 10 and 1000 times faster than the other methods.}
\label{fig:svm_openml}
\end{figure*}

\begin{figure*}[t]
 \centering
 \begin{subfigure}
  \centering
  \includegraphics[width=0.49\textwidth]{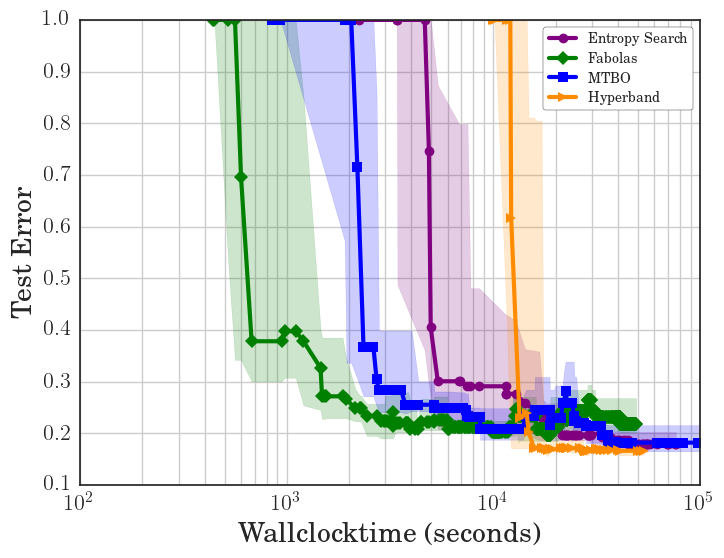}
  \includegraphics[width=0.49\textwidth]{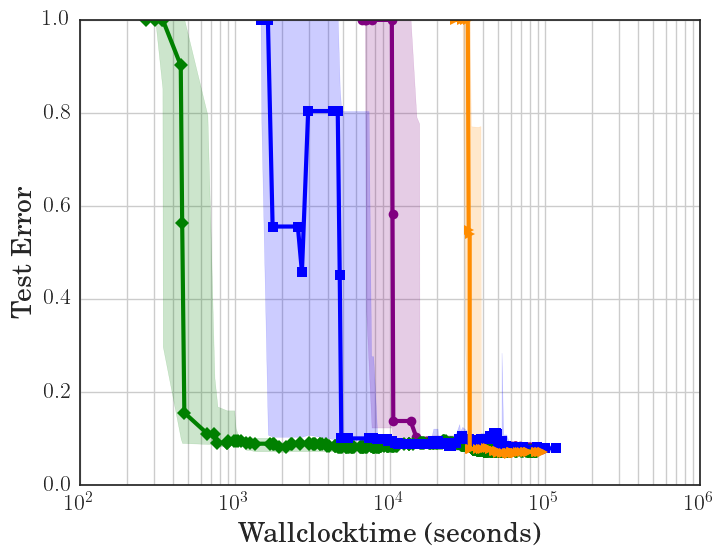}
 \end{subfigure}
\vspace*{-0.2cm}
\caption{Test performance of a convolutional neural network on CIFAR10 (left) and SVHN (right). }
\label{fig:cnn}
\vspace*{-0.2cm}
\end{figure*}

\subsection{Support vector machines on various datasets}

For a more realistic scenario, we optimized the same SVM hyperparameters without a grid constraint on MNIST and two other prominent UCI datasets (gathered from OpenML \citep{vanschoren-sigkdd13a}), vehicle registration~\citep{turing1987vehicle} and forest cover types~\citep{blackard1999comparative} with more than 50000 data points, now also comparing to Hyperband. Training SVMs on these datasets can take several hours, and Figure \ref{fig:svm_openml} shows that \fabolas{} found good configurations for them between 10 and 1000 times faster than the other methods.

Hyperband required a relatively long time until it recommended its first hyperparameter setting, but this first recommendation was already very good, making Hyperband substantially faster to find good settings than standard Bayesian optimization running on the full dataset. However, \fabolas{} typically returned configurations with the same quality another order of magnitude faster.

\subsection{Convolutional neural networks on CIFAR-10 and SVHN}

Convolutional neural networks (CNNs) have shown superior performance on a variety of computer vision and speech recognition benchmarks, but finding good hyperparameter settings remains challenging, and almost no theoretical guarantees exist.
Tuning CNNs for modern, large datasets is often infeasible via standard Bayesian optimization; in fact, this motivated the development of \fabolas{}.

We experimented with hyperparameter optimization for CNNs on two well-established object recognition datasets, namely CIFAR10~\citep{krizhevsky-tech09a} and SVHN~\citep{SVHN}.
We used the same setup for both datasets (a CNN with three convolutional layers, with batch normalization~\citep{ioffe-icml15} in each layer, optimized using Adam~\citep{Kingma-arxiv14a}).
We considered a total of five hyperparameters: the initial learning rate, the batch size and the number of units in each layer.
For CIFAR10, we used 40000 images for training, 10000 to estimate validation error, and the standard 10000 hold-out images to estimate the final test performance of incumbents. 
For SVHN, we used 6000 of the 73257 training images to estimate validation error, the rest for training, and the standard 26032 images for testing.

The results in Figure \ref{fig:cnn} show that---compared to the SVM tasks---\fabolas{}' speedup was smaller because CNNs scale linearly in the number of datapoints. Nevertheless, it found good configurations about 10 times faster than vanilla Bayesian optimization. For the same reason of linear scaling, Hyperband was substantially slower than vanilla Bayesian optimization to make a recommendation, but it did find good hyperparameter settings when given enough time.

\subsection{Residual neural network on CIFAR-10}\label{sec:resnet_exp}

\begin{figure}[t]
 \includegraphics[width=\columnwidth]{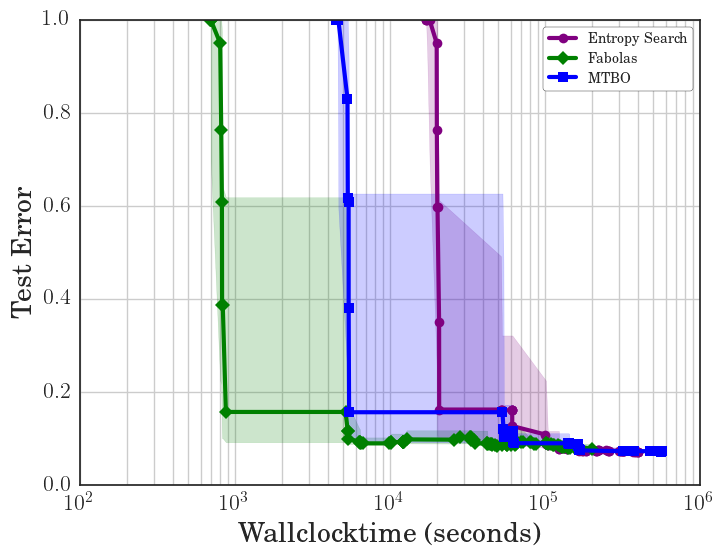}
 \caption{Validation performance of a residual network on CIFAR10.}
 \label{fig:res_nets}
\end{figure}

In the final experiment, we evaluated the performance of our method further on a more expensive benchmark, optimizing the validation performance of a deep residual network on the CIFAR10 dataset, using the original architecture from \cite{ResNets}.
As hyperparameters we exposed the learning rate, $L_2$ regularization, momentum and the factor by which the learning rate is multiplied after $41$ and $61$ epochs.

Figure \ref{fig:res_nets} shows that \fabolas{} found configurations with reasonable performance roughly 10 times faster than ES and MTBO. 
Note that due to limited computational capacities, we were unable to run Hyperband on this benchmark: a single iteration took longer than a day, making it prohibitively expensive. (Also note that by that time all other methods had already found good hyperparameter settings.)
We want to emphasize that the runtime could be improved by adapting Hyperband's parameters to the benchmark, but we decided to keep all methods' parameters fixed throughout the experiments to also show their robustness.

\section{Conclusion}\label{sec:conclusion}

We presented \fabolas{}, a new Bayesian optimization method based on entropy search that mimics human experts in evaluating algorithms on subsets of the data to quickly gather information about good hyperparameter settings.
\fabolas{} extends the standard way of modelling the objective function by treating the dataset size as an additional continuous input variable. This allows the incorporation of strong prior information.
It models the time it takes to evaluate a configuration and aims to evaluate points that yield---per time spent---the most information about the globally best hyperparameters for the full dataset.
In various hyperparameter optimization experiments using support vector machines and deep neural networks, \fabolas{} often found good configurations 10 to 100 times faster than the related approach of Multi-Task Bayesian optimization, Hyperband and standard Bayesian optimization. Our open-source code is available at \href{https://github.com/automl/RoBO}{https://github.com/automl/RoBO}, along with scripts for reproducing our experiments.

In future work, we plan to expand our algorithm to model other environmental variables, such as the resolution size of images, the number of classes, and the number of epochs, and we expect this to yield additional speedups.
Since our method reduces the cost of individual function evaluations but requires more of these cheaper evaluations, we expect the cubic complexity of Gaussian processes to become the limiting factor in many practical applications. We therefore plan to extend this work to other model classes, such as Bayesian neural networks~\citep{neal-1995, lobato-15,blundell-2015,springenberg-nips2016,klein-iclr17}, which may lower the computational overhead while having similar predictive quality.

%

\note{SF: References need some more attention, First names should always be abbreviated, certain letters should be capatilized, etc\dots}

\setlength{\bibsep}{0pt plus 0.3ex}
{\small
\bibliography{shortstrings,local,lib,shortproc}
\bibliographystyle{unsrtnat}
}

\clearpage
\appendix

\end{document}